# A Temporal Difference Reinforcement Learning Theory of Emotion: unifying emotion, cognition and adaptive behavior.

*Joost Broekens*
*Interactive Intelligence Group*
*Intelligent System Department*
*TU Delft, the Netherlands*

**Abstract**. Emotions are intimately tied to motivation and the adaptation of behavior, and many animal species show evidence of emotions in their behavior. Therefore, emotions must be related to powerful mechanisms that aid survival, and, emotions must be evolutionary continuous phenomena. How and why did emotions evolve in nature, how do events get emotionally appraised, how do emotions relate to cognitive complexity, and, how do they impact behavior and learning? In this article I propose that all emotions are manifestations of reward processing, in particular Temporal Difference (TD) error assessment. Reinforcement Learning (RL) is a powerful computational model for the learning of goal oriented tasks by exploration and feedback. Evidence indicates that RL-like processes exist in many animal species. Key in the processing of feedback in RL is the notion of TD error, the assessment of how much better or worse a situation just became, compared to what was previously expected (or, the estimated gain or loss of utility – or well-being – resulting from new evidence). I propose a *TDRL Theory of Emotion* and discuss its ramifications for our understanding of emotions in humans, animals and machines, and present psychological, neurobiological and computational evidence in its support.

**Introduction**

Emotions are essential in development. First, in infant-parent learning settings a child's expression of emotion is critical for an observer's understanding of the state of the child in the context of the learning process (Buss & Kiel, 2004). Second, emotional expressions of parents are critical feedback signals to children providing children with an evaluative reference of what just happened (Chong, Werker, Russell, & Carroll, 2003; Klinnert, 1984). In general, emotions plays a key role in shaping human behaviour. On an interpersonal level, emotions have a communicative function: the expression of an emotion can be used by others as a social feedback signal as well as a means to empathize with the expresser (Fischer & Manstead, 2008). On an intrapersonal level emotions have a reflective function (Oatley, 2010) and motivational function (Frijda, 2004): emotions shape behavior by providing feedback on past, current and future situations (Baumeister, Vohs, & Nathan DeWall, 2007) and emotions influence action tendencies. Finally, emotions result from event-related appraisal of personal relevance (Moors, Ellsworth, Scherer, & Frijda, 2013). In this article I define emotions as *valenced experiences in reaction to (mental) events providing feedback to modify future action tendencies, grounded in bodily and homeostatic sensations and evolutionary developed serviceable habits*. In Section 2 I will present an elaborate argument towards this definition.

Human emotions are complex phenomena. Consider the emotion of guilt. It requires many information processing abilities to be fully developed including the ability to imagine yourself in someone else's shoes, the ability to recognize that there is something like "others" in the first place, needing the ability to recognize the self. Further, you need to realize that you are responsible for, or at the very least causally connected to, the harm of the other. Finally you need to assess harm, presumably grounded in your own body.

Human adults are fully developed beings with a wide pallet of emotions. This includes the classical basic emotions that are reliably expressed on faces such as frustration, sadness, joy and fear, but also includes emotions that are typically considered social, such as guilt, shame, reproach, envy, and schadenfreude (gloating). However, one can easily expand this list with feelings of which it is not clear if they should be considered emotions, such as rebellion (liking the fact that you don't obey), awe (amazement about the unreal beauty or specialness of something or someone), and anxiety (the constant feeling of danger resulting in alertness and associated stress). These are definitely clearly identifiable feelings but are they emotions? Feelings typically refer to the conscious representation of the emotional state, e.g. as proposed by Damasio (Damasio, 1994) and LeDoux (LeDoux, 1996). However, this is not the distinction I use here. I use the term feeling to refer to any thought or stream of thoughts with associated positive or negative affective content. Some feelings are clearly distinguishable from emotions, such as feeling sick. Other feelings are clearly emotional, such as feeling joy. Human adult mental life is full of feelings, but is it full of emotions?

Young children are simpler beings. Babies do not seem to differentiate between different positive emotions (undifferentiated pleasure) and between different negative emotions at all (undifferentiated distress) (Sroufe, 1997). Infants frequently experience frustration, sadness, joy, and fear, but seldom experience guilt and shame. Even when children are capable of experiencing guilt and shame, such emotions are still rare compared to the more "basic" emotions. On top of that, infants and children express and experience emotions more frequently and more intensely than adults. Emotions are expressed all the time during play and interaction with others, children cry, laugh and get angry more often and more intensely. Is that the result of prefrontal cortical inhibition or is that the result of children not having lived as long as adults and therefore their life *is* more emotionally salient because they spent less time forming habits?

Many higher animals show evidence of emotions in their behavior. Of course we think of chimps, bonobos and dolphins. However, rats, mice, dogs, cats, bears and many other mammals and birds show clear signs of rich emotional life (Bekoff, 2008). And why would they not? They are living, surviving animals. More importantly, all mammals and the majority of birds are social by nature: they raise kids.

Many lower animals show emotional behavior (J. Panksepp, 1982). Reptiles show signs of rage, sexual drive, fear, and relaxation. What kind of emotions are these? Perhaps proto-emotions, or biological drives? What about fish? They startle, forage and mate. Are those proto-emotion, drives, behaviors? What is the distinction between these?

Unless we have a precise definition of emotion, one that is more precise, more mechanism-oriented and better testable than the definition I gave in the introductory paragraph, there is no way to answer these questions. There is unclarity about what emotions are both at the "higher human adult end" of the emotional spectrum as well as on the "lower reptilian end".

In this article I propose a simple hypothesis of what emotion is at its core. I propose, building on initial ideas by Brown and Wagner (Brown & Wagner, 1964) and Redish (Redish, 2004; Redish, Jensen, Johnson, & Kurth-Nelson, 2007), and extending ideas of Baumeister (Baumeister et al., 2007) and Rolls (Rolls, 2000), and work on intrinsic motivation (Singh, Lewis, Barto, & Sorg, 2010) that all emotions are manifestations of reward processing in Reinforcement Learning, *in particular that all emotions are manifestations of neural temporal difference assessment*. I show that the emotional episode shares its essential elements – event triggered, feedback providing, action tendency related, valenced experience, and bodily - with the assessment of temporal difference errors. I refer to this as the *Temporal Difference Reinforcement Learning (TDRL) Theory of Emotion.* This theory has limits, as the final discussion will highlight. It does however provide a strict base definition of emotion for further scrutiny and it provides clear boundaries for what an emotion is and is not.

In the rest of this article I will explain this hypothesis, show experimental, theoretical, and *in computo* support for it. Finally I examine the boundaries of this hypothesis. However, I first give an intuitive example of how the TDRL view explains fear as a manifestation of temporal difference processing.

**TDRL Emotion Theory in a nutshell**

Reinforcement Learning (RL) is a powerful computational model for the learning of goal oriented tasks by exploration and feedback (Sutton & Barto, 1998). Evidence indicates that RL-like processes exist in many animal species (Suri, 2002). Key in the processing of feedback in RL is the notion of TD error, the assessment of how much better or worse a situation just became, compared to what was previously expected.

In the TDRL Theory of emotion, distress (or joy when positive) is the manifestation of the assessment that your situation just got worse (or better). Worse means a decrease in value, which is a decrease in a common motivational currency as explained in later sections. Therefore, distress is the manifestation of *negative adjustment, which is in RL terms equivalent to a negative temporal difference error*.

Fear (or hope when positive) in the TDRL Theory of emotion is the anticipation of distress (or joy when positive). Fear is thus the manifestation of anticipation of worsening of your situation. By anticipation I mean the imagining of a possible future state. Fear is thus the anticipated experienced distress where one to actually "move" towards and arrive in that state[1]. As distress is a relative signal, fear is also a relative signal. Let's assume you imagine a trace of interactions ending in some state, *s*. If the value of that imagined state, *s*, has already been completely incorporated in the trace towards state *s* (let's say a habit has formed), then all adjustments have been made already. In other words, you will not feel fear in that case. If, however, you experience negative adjustments during the imagination of the trace towards state *s*, then fear will be felt. Fear is thus a special case of distress, namely, fear is *the manifestation of anticipation of negative adjustment*. This assumes you can look ahead, i.e., you have a model of the world allowing you to anticipate negative adjustment (fear) in addition to calculate actual negative adjustment (distress). Fear therefore is very close to distress. This mechanism of anticipation is also the basic driver for subsequent regret and disappointment, in line with Reisenzein (Reisenzein, 2009b) and Mellers (Mellers, Schwartz, Ho, & Ritov, 1997), to which I will come later.

Let's take the example of my son playing and climbing rocks (Figure 1, left). My son plays and has fun and does not feel fear. Now consider the second setting (Figure 1, right): my son looking into a deep ravine, with my wife holding my son. My wife feels fear. However, the first setting is objectively more dangerous than the second. How come my son does not feel fear in either of these settings, while my wife feels strong fear even in the second setting? This example illustrates two important things: fear is relative and fear depends on anticipation, i.e., a model that allows lookahead. My son, being a child aged 6 at that time, does not anticipate distress due to slipping and falling. He lives in the now. He does not experience any (anticipated) negative adjustment. My wife, being an adult, does anticipate negative adjustment, because (a) she can imagine that children do random things and (b) she experiences imagined distress due to her son falling off a ravine because this needs significant negative adjustment to her situation. This generates fear.

---

[1] Note that by "move" and "arrive" I do not literally mean movement but more so get closer in the abstract sense of the word, i.e., it is sufficient to imagine that there is a tendency towards ending up in that future state, potentially but not necessarily due to your own behavior.

The TDRL Theory of Emotion interprets this scenario as follows. My wife simulates state transitions that involve major negative adjustments (the death of our child by falling into the ravine[2]). These thoughts generate negative adjustments, of which my wife knows they are not actually happening, but still they are felt as real adjustments. These felt adjustments are actual temporal difference errors. This is fear: internally simulated negative temporal difference errors. If the event would happen as such, it would cause distress, i.e. actual negative temporal difference (TD) errors without the luxury of simulating an alternative future. The feeling of fear thus needs (a) the ability to feel distress, which I propose *is* a negative TD error, (b) the ability to anticipate[3], and (c) the triggering of distress (TD errors) while anticipating. This also explains why fear "appears to be" such a powerful motivator. I write "appears" because in the TDRL Theory of Emotion fear *is* the manifestation of anticipated negative adjustment. As this anticipation of negative adjustment must be assessed using similar neural mechanisms as actual negative adjustment, it enables the individual to change future action motivation and subsequently chose behavior that reflects this adjustment. It is not the fear that drives the behavior, fear is the manifestation of anticipated negative adjustment and this adjustment leads to a different behavioral outcome in the now. My wife anticipates negative adjustment *and* feels fear, then holds our son firmly.

This interpretation of fear and distress helps us understand many fear-related phenomena. Children are fearless and worry-free compared to adults because they have less elaborate anticipation capabilities. They have less habits, less cognitive control, working memory capacity, and top-down attention typically attributed to their underdeveloped inhibitory pre-frontal cortex capacity. Babies show signs of distress but not fear, as they have not yet developed any form of anticipation of interactions with the world. Rumination (e.g. in depression) causes actual distress because continuous activation of negative TD error calculations is as if these negative adjustments are happening. Positive rumination is an effective treatment for depression, for exactly the same reason. Depression relates to inactivity, if all options seem to involve negative adjustment, then why do anything in the first place? These and other phenomena will be dealt with in more detail later in this article.

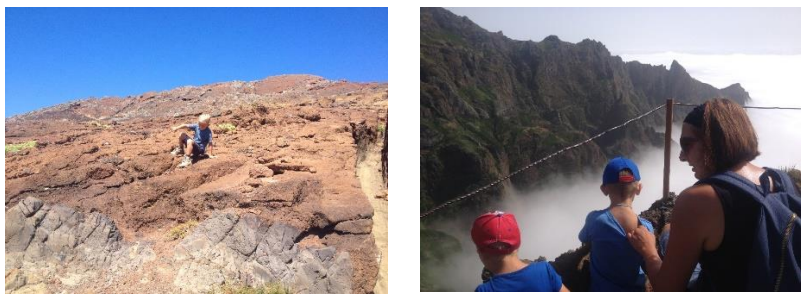

*Figure 1. Fear as anticipation of negative adjustment*

---

[2] My children are doing fine by the way.

[3] which *is* model-based lookup in Reinforcement Learning terms. In this case, the individual learns a model of the action-environment statistics $P(s'|s,a)$, i.e., how do states change as a result of actions in particular states.

**Emotion**

*What I consider to be emotions*

To systematically relate emotions to TD error processing, I have to precisely define what I consider emotions and what not. I propose 5 essential elements of emotion. First, I consider emotions to be *reactions to external or internal events*. An event is a measurable change in the state of the organism. It does not matter if this change is represented symbolically or not. What matters is that the organism is able to deduce that the situation changed with respect to just before. An event is thus a *state change*. Whether this is an imagined state change (a thought) or an actual state change does not matter either. This is in line with embodied cognition ideas of Cotterill (Cotterill, 2001), Heslow (Hesslow, 2002) and Ziemke (Ziemke, Jirenhed, & Hesslow, 2005) proposing that a thought is grounded in simulation of actual behavior. Thoughts are internal simulations of behavior. For example, *you* just before imagining winning the lottery are different from *you* just after imagining winning the lottery. It does not matter that this though is not true, it is a state change anyway. This event-relatedness aspect of emotions also excludes longer-term affective states such as depression and mood (Beedie, Terry, & Lane, 2005) from emotions. It generalizes the often-seen object/person-relatedness as a common aspect of emotion in most cognitive emotion theories (Moors et al., 2013; Ortony, Clore, & Collins, 1988): an emotion is related to a state change.

Second, emotions are *valenced experiences* (Moors et al., 2013; Ortony et al., 1988). Valenced means that there is an assessment of positiveness versus negativeness based on personal relevance. As emotions are reactions to events, this means that emotions always involve an assessment of positiveness versus negativeness of a *state change*. For example, if you learn that your flight home is canceled due to severe weather you will assess this as negative, because it impairs achieving your house which is a desired state. Feelings that relate to the general state of information processing are not considered emotions. For example, surprise, confusion and flow are usually considered affective feelings and some even refer to these feelings as emotions. However, such feelings are more related to the general state of information processing, rather than event-related assessments of positiveness and negativeness. Surprise does not involve valence. It involves the realization of some form of novelty. Confusion might involve a positive or negative feeling but not necessarily so. Puzzlement can be both positive as well as negative, and additionally again relates to general information processing (e.g., inconsistent information). Flow is typically positive, but not an event related phenomenon. Information about information processing can be involved in the elicitation of emotion, such as in the case of novelty detection, but is not a sufficient condition for emotion.

Third, emotions are *not equivalent to but grounded in bodily or homeostatic sensations*. In this view not all feelings[4] associated with positiveness or negativeness are emotions. There are many feelings that involve detection of harm and benefit either based on direct sensations or homeostatic processes. For example, pain is an unpleasant feeling, but not an emotion. Pain is always pain, it never becomes nausea or the perception of sucrose. Pain, just like sugar, umami, bitter, gentle touch, hunger, thirst, cold, hot, and many more, are hardwired signals with inherent meaning, often functioning as drives for specific behaviors. They arise consistently, and their meaning is not changeable. This is also the reason why I do not consider gustatory disgust an emotion. This form of disgust is the gustatory reaction to something foul tasting or unknown in the mouth focused at spitting out the thing that causes the sensation and preparing for throwing up. It is an inherent reaction and it is not changeable. Sensations such as pain, disgust and gustatory satisfaction are multimodal components of the evaluation of pleasantness (in RL terms the reward), but they are not emotions proper; emotions are about pleasure and displeasure which is more like a common currency (Cabanac, 1992).

---

[4] The term feeling is used in the sense explained in the introduction.

Forth, strongly related to Baumeister et al (Baumeister et al., 2007), I consider all emotions to be *feedback signals in some form*. Baumeister identifies feedback, anticipation and reflection as different signals, I consider these to be different types of feedback signals. In their simplest form emotions are feedback signals about the current event: how much did this event make my world better or worse? This is the case for joy and distress (I use joy and distress loosely as terms for the evaluation of positiveness/negativeness of events). In more complex forms emotions are feedback signals about possible future events. How desirable is a particular future event? This is the case for hope and fear (I use hope and fear loosely for the positiveness / negativeness of a potential future event). This is anticipatory feedback, the type of feedback that, when it impacts action selection, consistently biases behavior into a particular direction. Even more complex emotions are feedback signals reflecting upon a situation. This is the case for, e.g., disappointment and relief. Disappointment and relief give feedback about the correctness of the hope and fear signal (Reisenzein, 2009b). In essence they are corrections (reflective feedback) on anticipated joy and distress. The TDRL Theory of emotion is able to distinguish clearly between joy, relief, and hope, as well as distress, fear, and disappointment (see later).

Fifth, emotions are *manifestations of future behavioral change*, in line with Frijda's views on emotion, motivation and action tendencies (Frijda, 2004). Emotions motivate action in particular directions, because emotion is a manifestation of (future) behavioral adjustment. However, a particular emotion does not necessarily motivate a particular behavior. Emotional valence and intensity thereof drive the motivational power of an emotion, consistent with emotion as approach versus avoidance signal.

These five criteria result in the explicit exclusion of several emotions that have been considered basic. For example, rage-aggression, horny-sexual drive, fear-startle, and surprise are not considered emotions proper, even though they are definitely *serviceable habits with an associated feeling*. They are behavioral responses to threatening stimuli or internal drives and they serve survival. However, they are not event-related evaluation of behavioral adjustment. All three are strong behavioral motivators but they are evolutionary hard-wired processes not in need of a valence assessment of an event. Surprise is not related to valence assessment either (see e.g. Reisenzein). Further, rage and sexual drive last longer than typical emotions and are not necessarily related to future behavioral change, but at resolving a current biological need. Further, primitive animals show these behaviors, including flies, spiders and fish. I define such phenomena as proto-emotions for lack of a better term, but not as emotions proper.

To wrap up this section, I define emotions as *valenced experiences in reaction to (mental) events providing feedback to modify future action tendencies, grounded in bodily and homeostatic sensations and evolutionary developed serviceable habits*.

*Emotions and evolution.*

Emotions are a product of life with central nervous systems. So to answer the question why and how emotions evolved in nature I first address this question for brains in general. Nervous cells evolved to enable evolutionary preprogramming of reflexes for multicellular organisms. This functionality is needed to control the different parts of a multicellular body in response to external and internal stimuli. Central nervous systems evolved to do the same but in a more coordinated fashion for more complex bodies. If the body needs to respond as a whole, rather than as separate parts, you need a central nervous system with fast, myelinated, axons to coordinate that response (Susuki, 2010). As such it is safe to say that simple brains evolved to be able to evolutionary encode complex reflexes. The pain response, low sugar levels that trigger foraging behavior, and reproductive signals of the other sex triggering reproductive behavior are essentially complex reflex mechanisms that serve survival.

The next step in the evolution of brains is to not only *encode* reflexes but also *learn* the appropriateness of reflexes. To do this, central nervous systems need two properties. First, cells need to be able to encode covariance (e.g., through Hebbian learning). Second, cells need to be able to inhibit rather than only excite other cells. With these two properties, learning is a matter of encoding co-occurrence of sensorimotor activity, if the nervous system is pre-wired in the right way. For example, if you are a small frog and swallow a green and juicy looking, but foul tasting insect, then the coactivity between nervous cells responsible for the biting, the sensing of the green insect, and the sensing of the foul taste (inhibitory neurons) is enough to reduce the probability that you will bite that insect the next time (Fig 2). If all non-green insects taste great, than you will learn to avoid biting green insects only.

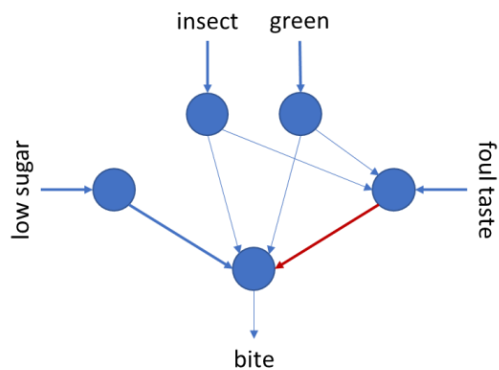

*Figure 2. A simple neural system for the learning and adaption of reflexes. Thick arrows represent fixed non-adaptable connections. The red think arrow represents a fixed non-adaptable inhibitory connection. The thin arrows represent Hebbian connections. Low sugar levels excite biting. Foul taste inhibits biting. If biting in green insects consistently activates the foul taste detector, but biting in all other insects does not, than the result after repeated exposure is a strong excitation of the foul taste detector when green is detected, resulting in inhibition of biting. This effectively adapts the low sugar -> eating reflex. Of course this network needs careful setting of thresholds etc., and evolution had about 1000 million years to do so.*

Three essential properties of central nervous systems that allow further flexibilization of reflexes towards proper task learning include: dealing with delayed effects, a common currency, and play. First, some effects of behavior are delayed. The simple reflex adaptation network presented above cannot deal with such delayed effects. If you would become ill after 15 minutes, then the neurons encoding for green and biting are not active anymore, and the association between foul tasting and biting green stuff cannot be built. This severely limits survival potential of the organism. Dealing with delayed effects is the next big step in learning the appropriateness of reflexes.

Second, there is a large variety of positive and negative effects (pain, hunger, thirst, gentle touch, sugary substances, a full belly). In order to learn tradeoffs between different, competing, behaviors it is essential to be able to weigh these effects. An organisms needs a common currency to decide whether foul tasting food is worse than starvation, or, whether being hurt is worth the effort of hunting a nice meal (Cabanac, 1992). I refer to this as the *multimodality* of positive and negative effects of behavior. This does not mean that such common currency needs to be encoded in the brain somewhere separately as a prerequisite for action selection, neither that the currency is some abstraction over the different sources of good and bad, but rather that a *multimodal representation of value* must exist as a consequence of an organism's ability to deal with a variety of "good and bad" at behavior selection time. At some level in the neural command chain, a behavioral preference is built for hunting versus waiting, and that behavioral preference *is a proxy for a common currency*. Once this proxy exists, it can be used as basis for other neural calculations, such as the reward and value/utility used in TD learning. Optimizing pleasure and minimizing displeasure is therefore a plausible thing to try for advanced living species (Cabanac, 1992).

Third, an organism needs to show behavioral flexibility (Bekoff, 2008). To enable this, an organism needs to generate novel behaviors and repeatedly test the effects thereof rather than to only adapt pre-existing ones. If an animal will only try to bite when hungry, then it will never learn to bite in other situations, unless evolution has given it a pre-wired exciting connection between another drive and biting (e.g., aggression when facing rivals). This is fine for simple environments, but not for complex ones in which survival depends on the successful learning of complex tasks. It is impractical (and complexity-wise impossible) to evolutionary encode all possible reflexes that are needed to, for example, hunt for ants in a tree trunk with a thin stick (chimps), throw nuts before cars at traffic lights and then eat the nut when the lights turn red (corvids), and to drive a car through busy traffic (humans). Therefore, to learn novel tasks, organisms need to try out new behavior preferably in a safe environment. This is called play. In nature, play is strongly related to child raising for the simple reason that play needs to be safe otherwise it kills you. Play is the behavioral expression of exploration. Exploration is essential for the rewiring of existing and the learning of new reflexes. Parents are needed to shape the behavior (correct and motivate) as well as to restrict the environment (create the proverbial sandbox). Parenting and play must have co-evolved. Curiosity did not kill the cat, because it had parents that shaped the sandbox.

To adapt existing reflexes and learn novel reflexes with delayed multimodal effects, an organism needs a common currency to evaluate what to do next, needs to associate this currency to interaction with the environment in the past, and needs to play in a safe environment. As we will see later, these three properties of organisms are represented in the computational TDRL learning method. However, here we focus on emotion.

In this adaptive behavior stance towards evolutionary brain development, the question of how and why emotions developed almost becomes obsolete. To do all this, emotions are really not needed. However, as emotions are definitely observed in many animals that adapt existing reflexes and learn novel reflexes with delayed multimodal effects (such as birds and mammals), there must be a link. I propose that emotions are the manifestation of mechanisms that aid an organism to adapt and learn novel complex reflexes. Therefore there is a strong link between emotions and survival-rated reflexes such as sexual drive, aggression, and pain responses. These are very powerful and important reflexes involving strong drives and strong effects. The strongest and most primitive emotions must therefore be grounded in attempts to adapt such reflexes. However, the reflex is not an emotion. It is perhaps the behavioral correlate of the emotion but only in those cases where the reflex itself is triggered too. Further, emotion is related to valuing interaction with the environment. To do this, the notion of a common currency had to evolve. At least one animal even invented ways to externalize this common currency, i.e., homo sapiens' concept of money. It makes sense to assume that evolution "invented" this concept much earlier as a way to bargain between alternatives of action and to store future potential of behavior. The ability to value is not the same as the ability to adapt value. The first refers to dealing with multimodality of effects in order to solve an action selection problem, the latter refers to increasing or decreasing the magnitude of the multimodal representation of effects associated with a situation. I propose that emotions get valence from the latter process. Put strictly, valence and intensity of an emotional episode is the sign and magnitude of the adaptation of the multimodal representation of effects associated with a situation.

To conclude, this evolutionary and adaptive behavior view proposes that emotions are a natural consequence of ever more intricate adaptation of reflexes shaping complex behavioral responses to help survival of the individual. In later sections I explain how the TDRL perspective can shed light on what individual emotions are in light of this adaptation.

*Emotions and cognitive complexity.*

Emotions are intimately tied to cognitive complexity. More cognitively complex individuals and species have more complex and more variety of emotion. Observations from developmental psychology show that children start with undifferentiated distress and joy, growing up to be individuals with emotions including guilt, reproach, pride, and relief, all of which need significant cognitive abilities to be in developed. In the first months of infancy, children exhibit a narrow range of emotions, consisting of distress and pleasure (Sroufe, 1997). Joy and sadness emerge by 3 months, anger around 4 to 6 months with fear usually reported first at 7 or 8 months (Sroufe, 1997). To be able to feel anger, one needs a concept of self and other. To be able to feel relief, one needs the ability to anticipate distress and compare the anticipated distress with the actual distress. These abilities develop over time. More complex emotions accompany this cognitive development.

Cognitive appraisal theory follows the same lines. Appraisals are the evaluative components associated with particular emotions. As proposed by Scherer, appraisals exist at different levels of complexity, and simpler levels are available earlier in individual development (Scherer, 2001). A baby has no way of detecting agency and responsibility and therefore cannot be angry at its caregivers. It can feel distress about not getting what it needs, but this is not the same as being angry. A toddler aged 1½ years has not yet developed a solid theory of mind of others. Therefore it cannot feel guilt, envy, reproach, happy-for and sad-for another person. Young children will take away toys from other children because they do not realize that the other experiences "a decrease in perceived multimodal value". In appraisal terms they lack the appraisal of agency and cannot appraise whether something is beneficial or not for someone else (Ortony et al., 1988).

To conclude, the complexity of emotions depends on the cognitive complexity. This holds for individual development (this section) and evolutionary development (previous section). I propose that there is some level of cognitive complexity (which I, in lack of a better term, equate to the complexity of the central nervous system) that is required before one can speak of emotions. I define this *necessary and sufficient* level as the ability to process delayed multimodal effects aimed at adapting existing, and learning novel reflexes. Behavioral signs of the existence of emotions in organisms include play, child raising behavior, conflict resolution, and the ability to learn novel tasks that involve delayed rewards. Humans share these behaviors with a vast number of other animals, raising serious doubt about the often human-centered view of what emotions are and who can or cannot experience emotions. I discuss the ramifications of this view in light of animal welfare later in this article.

**Reinforcement Learning**

*Sequential Decision Making, Reinforcement Learning and Instrumental Conditioning ++*

In this article I argue that Temporal Difference Reinforcement Learning (TDRL) is a suitable model system for understanding the relation between emotion, cognition and task learning. Reinforcement Learning (RL) is a powerful computational model for the learning of goal oriented tasks by exploration and delayed reward (Sutton & Barto, 1998; Tesauro, 1995; Watkins, 1989). It can be described as "instrumental conditioning ++". The key difference is that RL can learn situational action values based on delayed reward signals received after a trace of actions, whereas instrumental conditioning only deals with immediate reward/punishment. In other words, RL propagates back rewards through time in an attempt to solve what is called the credit assignment problem: how much credit should each action in the trace get based on the rewards received in that trace. RL solves a *sequential decision making problem* by learning action values for sequences of actions based on the delayed rewards. An *action value* in RL represents the immediate and expected future payoff of the action and is recursively defined and situation dependent. The *state* in RL refers to the current situation as perceived by the animal. The *reward* in RL is defined as the immediate positive (encouragement) or negative (punishment) feedback signals determining what should be sought after by the animal. It refers to the common currency mentioned above. The animal is usually referred to as the *agent* in RL. I use the term agent in this article as a reference to either animal (natural agent) or autonomous robot (artificial physical agent) or simulated robot / virtual character (artificial virtual agent). RL has been used successfully to enable agents to learn a variety of tasks (Kober, Bagnell, & Peters, 2013) as well, with the right function approximator, solve control problems in many other complex domains (LeCun, Bengio, & Hinton, 2015).

In practice, RL agents learn action values by exploration and when certain enough or after a given amount of exploration effort the agent then uses the action values in its action selection mechanism to select appropriate actions in states. An important distinction in RL is thus the learning of the values versus the use of the values in action selection. These are separate processes. The action selection mechanisms typically gives more chance to higher valued (or underexplored) actions. As action values are learned dependent on the state, in each state an agent can learn to select different actions as long as there are some discriminative features of that state that make it different enough from others[5]. This effectively cause the agent to move from one state (situation) to another by sequentially selecting actions[6].

---

[5] This relates to fully observability in RL. For RL to work in practice, partial observability is sufficient in many cases.
[6] In nature this process is much less discrete and the transition between states and even between actions and stimuli is often continuous.

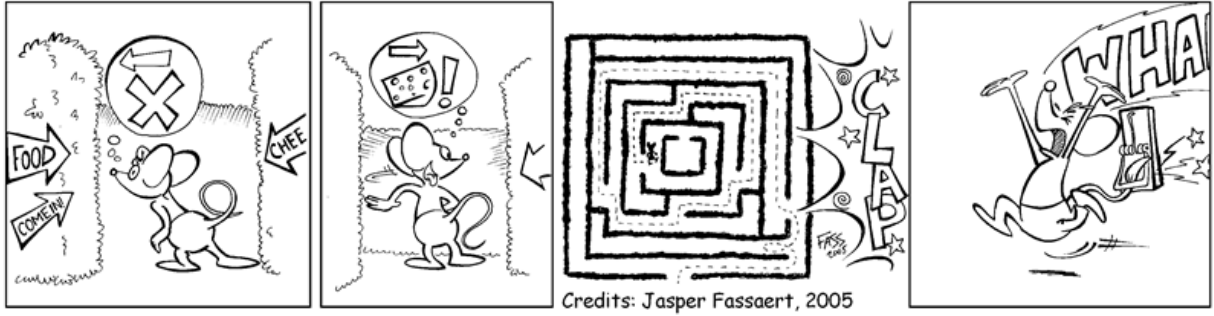

*Figure 3. A Cartoon representation of Reinforcement Learning*

*Delayed reward and the Temporal Difference signal*

Virtually all RL methods learn by updating action values based on the *temporal difference error*, a learning signal informing the agent how to adjust the value of its last action based on current and anticipated future rewards[7]. For the TDRL Theory of Emotion it suffices to assume the simplest form of action value learning that is based on TDRL, called SARSA (state-action-reward-state-action) (Rummery & Niranjan, 1994). It is representative of most other model-free forms of action value learning and is the easiest to explain when it comes to the dynamics of the *temporal difference* error.

$$Q(s_t, a_t) \leftarrow Q(s_t, a_t) + \alpha[r_t + \gamma Q(s_{t+1}, a_{t+1}) - Q(s_t, a_t)]$$

[1]

In SARSA (Figure 4, and Formula 1), the value $Q(s_t, a_t)$ of an action $a_t$ in a state $s_t$ at a certain moment in time $t$ is updated every time the agent visits state $s_t$ and executes action $a_t$. The magnitude of the update depends on the reward $r_t$ received at that moment in state $s_t$ after execution of action $a_t$ and the value $Q(s_{t+1}, a_{t+1})$ for the action that is selected in the next state. The signal that drives the change in the current value $Q(s_t, a_t)$ is expressed by the term $[r_t + \gamma Q(s_{t+1}, a_{t+1}) - Q(s_t, a_t)]$ that is weighted by learning rate $\alpha$. This term expresses the difference between the current value of action $a_t$ in state $s_t$ and the received reward $r_t$ in $s_t$ plus the discounted value of the next action $a_{t+1}$ in the next state $s_{t+1}$. In words this is the update that is needed to reflect that $Q(s_t, a_t)$ needs to encode for the current and all (discounted) future rewards assuming the actions typically taken by the agent. This update is the *temporal difference error* (encircled in Formula 1): the error in the prediction of the current $Q(s_t, a_t)$ given a newly received reward $r_t$ and the (potentially earlier updated) estimate of $Q(s_{t+1}, a_{t+1})$. This error signal is used to drive learning while the agent is allowed to sample many state transitions allowing action values to converge to the cumulative expected return of actions[8].

---

[7] Although RL methods that do not use sampling to learn values but learn or use a state transition model, i.e., model-based approaches, typically do not explicitly define the TD error, it still exist and can be calculated in a similar way for model-based value updates.

[8] Assuming a particular policy, a Markovian world and a small enough learning rate. These assumption are important for learning but not important for our main argument for the link between emotion and RL.

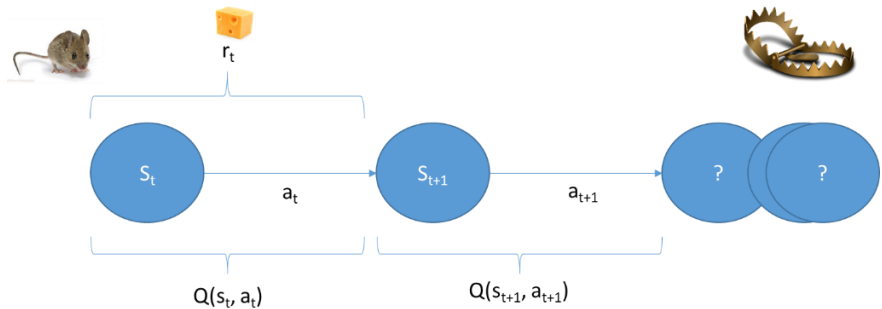

*Figure 4, Graphical representation of SARSA with an immediate reward and a delayed punishment. SARSA updates only when visiting the states (one step update). As such it takes some repetition before delayed signals are propagated back into the Q(s, a) of preceding actions.*

*Neurobiological and psychological evidence for TDRL.*

Apart from the computational benefits of TDRL as an algorithm to learn tasks based on delayed reward, such as robot motor control and complex decisioning problems, TDRL has been argued to be a biologically plausible mechanism for animal task learning (Hare, O'Doherty, Camerer, Schultz, & Rangel, 2008; Rolls, 2014; Schultz, Dayan, & Montague, 1997). Different neurons seem to encode for the temporal difference error, reward, and action values. Further, in psychology, RL has been proposed as a model for understanding decision making, for example in the domain of addiction (Redish, 2004; Redish et al., 2007), and it has been proposed as an alternative explanation for the earlier Iowa Gambling Task findings (Dunn, Dalgleish, & Lawrence, 2006; Maia & McClelland, 2004). Value-based action selection has been proposed to occur in the basal ganglia, in concert with pre-frontal cortex reward and value calculation (Bogacz & Gurney, 2007; Houk et al., 2007; Tanaka et al., 2004). These ideas are in line with the more recently discovered functions of the prefrontal cortex involving value and TD error calculations (Berridge, Robinson, & Aldridge, 2009; Haruno & Kawato, 2006; Suri, 2002; Tanaka et al., 2004), and the recently proposed role of the amygdala as a center for emotional value/valence processing (Murray, 2007). Finally, RL agents show several striking behavioral similarities with learning and playing animals[9]. Animals develop best when they are allowed to explore behavior in a relatively safe, parented environment still allowing for positive and negative feedback. These are essential biological conditions for the genesis of play (Burghardt, 2005). This is how RL agents are trained too, by trying out random actions in a virtual "sand-box" updating the action values based on feedback. Curiosity, i.e., the drive to try out new things, is absolutely necessary for RL agents as well, and a typical strategy to drive exploration in computational approaches that emphasize intrinsic motivation is to overvalue untried actions (Singh et al., 2010). Animals learn novel tasks by rewarding desired behavior not by punishing undesired behavior. Frequent punishment results in inactivity and hesitation, also seen in RL agents[10].

An exhaustive review of the biological and psychological evidence for reinforcement learning is out of scope for this article. However, it is important to highlight three major findings that seem to be quite consistent and important for our argument. First, RL-based reward processing seems to occur in the prefrontal cortex (with different areas involved in encoding of value versus TD error), amygdala and basal ganglia system. Second, these areas are traditionally involved in action selection and behavior. Third, these areas have been historically involved in emotion processing, motivation and coping (Bechara, Damasio, & Damasio, 2000; Damasio, 1994; LeDoux, 1996). It seems, therefore, that the areas that traditionally were believed to be involved in the processing of action selection and emotion, now also seem to be involved in RL processing. This is not surprising for action selection, as RL is about utility (or value)-based action selection. It is surprising that there is overlap between RL and emotion. This suggests that there is neurobiological evidence supporting the idea that emotion processing and reward processing in the RL sense are related processes in the brain.

---

[9] Interestingly, the original formulation of Q-learning was heavily inspired by animal learning theory (McNamara & Houston, 1985; Watkins, 1989).

[10] In instrumental conditioning, reward and punishment is in fact better understood as the TD error. The TD error is the feedback signal that makes an action more or less likely, and also depends on whether the reward was expected or not.

**Outline of the Argument**

The remainder of this article focuses on the argument that emotions are manifestations of temporal difference processing in reinforcement learning. The claim I make is straightforward and follows from the conclusions about how emotions evolved, what emotions do, and how TDRL learns.

1) The elicitation of emotion and the TD error is similar: emotions and the temporal difference error in reinforcement learning are feedback signals resulting from the evaluation of a particular current state or (imagined/remembered) event.
    a) Emotions are valenced reactions to (mental) events elicited by an assessment of the event's impact on personal relevance[11].
    b) The temporal difference error estimates how much better or worse the situation just became, based on currently received reward and predicted future reward.
2) The functional effect of emotion and the TD error is similar: emotions and the temporal difference error impact future behavior by influencing action motivation.
    a) Emotions impact future action selection either directly or by modifying action motivation.
    b) The temporal difference error updates action values.
3) The evolutionary purpose of emotions and the TD error is similar: both emotion and the TD error aim for long-term survival of the agent and the optimization of well-being.
    a) Emotions aid survival of the individual as a signal that can be used to optimizing well-being.
    b) The TD error is a signal that can be used to optimize long-term reward.

This brings us to the core of the argument summarized in the following infographic:

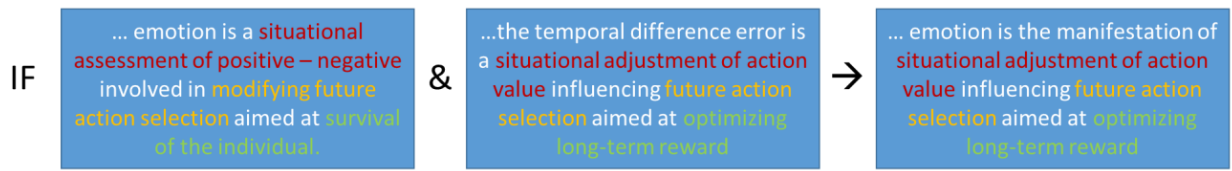

In later sections I will provide multidisciplinary evidence for this claim. In particular I will highlight computational, neurobiological and psychological support. First however I focus and clarify my claim.

---

[11] …and are grounded in bodily and homeostatic sensations and evolutionary developed serviceable habits.

*Framing and interpretation*

I propose Reinforcement Learning (RL) as a framework to study the grounding of emotion in evolution and development. To be precise, I intend to show that emotions are the manifestation of temporal difference errors. I show how this TDRL Theory of emotion provides important insights into our understanding of the positive and negative aspect of emotion, the event relatedness of emotion, the relation between emotion and other affective phenomena, the existence of emotion in other animals, and the relation between emotion and learning such as how more complex emotions emerge as a result of cognitive complexity.

In line with (Murray, 2007), it is important to realize that I do not claim that emotion and (an abstract representation of) reward/punishment are identical. Reward/punishment and emotion are not the same. First, reward processing is necessary for TD error processing but not equivalent. Reward in RL is a subjective[12] measure of goodness while the TD error is the adjustment of the action value[13]. This adjustment is derived from the reward signal, but also depends on an estimate of the value of the current situation (see section on RL above). So, the TD error signal is distinct from the reward signal. Second, even if an evaluation results in a common currency used to compare or update actions, this must be still built on top of the biological wetware of a body and what is good or bad *for* that body. Emotions are therefore not simply an abstract utility because this utility must have biological grounding. Of course at some point an adjustment needs to be made to make an action more or less likely to be selected in the future, and this needs to be a "coding of action motivation", but this code refers to something real and felt. To put it differently, action motivation is grounded in the multimodal physiology of reward and punishment (e.g., pain, smell, touch, disgust) as well as homeostatic processes that manage low-level urges and senses of well-being (e.g., sugar level, hunger, satisfaction) . At some point the action's motivational value is coded on a more abstract level (perhaps in the amygdala (Murray, 2007)), because a decision needs to be made, but the importance of the different reward modalities may very well change over time depending on the current physiological state and intrinsic motivational factors. This means that to properly understand the relation between emotion and *adaptive behavior in general*, one needs to study and simulate physiology because that is the basis for reward and punishment and hence the abstract encoding of value. Also in the TDRL Theory of Emotion, emotion is grounded in physiology, homeostasis, hormones, and the drive to survive.

I do not claim that *all* animal species have emotions. I claim that the attribution of emotion to animal species is linked to that species' ability to compute TD-like action value updates. If a brain lacks the processing capacity to do this, then it also lacks the ability to realize that the world got better or worse, and it definitely lacks the ability to anticipate adjustment. Instrumental conditioning might be sufficient for emotions of joy and distress, if one assumes that the TD error is what is referred to as reward in this process[14]. If the conditioning is realized with a relatively straightforward neural architecture that enables the conditioning of reflexes (see Figure 2), it might not be sufficient.

---

[12] The standard view is that reward is an objective measure form the environment but the intrinsic motivation literature in RL (Singh et al., 2010) proposes a more plausible subjective view on the reward signal, fully compatible with TDRL Emotion theory as even when using intrinsically motivated reward signals a TD error is calculated to update action values.

[13] In some formulations of RL the adjustment is a state value adjustment rather than an action value adjustment. This difference is not important for our discussion.

Relatedly, I *do* claim that many animal species have emotions. As soon as one can observe RL-like behavior (exploration, child care, task learning based on delayed rewards, conflict resolution, to name a few) it makes sense to talk about emotions proper. The set of animals with emotions therefore is huge.

Further, I propose an important role for social processes in emotions. I believe that social emotions are strongly grounded in RL. I will come back to this later, but in a nutshell: why would you care if someone is angry at you, and, why is guilt (a) a bad feeling and (b) a useful feedback signal to you? I believe there is a very simple reason that resonates with reinforcement learning: angry people communicate to you that they had to negatively adjust their situation due to something they think you had a hand in. Expression of anger is the direct social signal translation of that negative internal feedback signal, and, its function is to decrease the receiver's action tendency of the blameworthy action. In a very real sense, the expression of anger is the communication of a negative TD signal to the one to whom the expression is directed. If successfully communicated, guilt is *your* internal feedback signal reflecting the adjustment of *your motivation to repeat your offensive action*[15]. Showing guilt is again the direct social signal translation of that internal signal, intended to communicate that the message has arrived and incorporated in future action motivation. Many emotions have a social signal component and an internal evaluative component. The internal component always relates to the adjustment of action value, the TD error, in ever more complex situations. In social species it is beneficial to evolve the capability to process agency-related information in order to realize that someone had some form of action responsibility. It is also beneficial to evolve social signals to resolve conflict and avoid physical harm due to fighting. Non-social species probably lack emotions such as anger and guilt, not because they lack emotions altogether but because there was never any need to (a) process responsibility/agency and (b) communicate corrective signals. I would be comfortable claiming that dogs feel anger and guilt, but frogs do not.

Finally, I do not claim that machines that implement reinforcement learning to adapt their behavior feel anything. I claim that such machines simulate emotions. The simulation of a phenomenon is not the same as that phenomenon itself. A weather simulation does not create rain. A simulation of atoms forming molecules does not create actual chemistry. By simulating emotions I do not create feelings. Further, feeling is a term to be avoided in artificial beings because it refers to the experiential and machines cannot - at least at this point - experience anything in the naturalistic sense[16].

---

[14] Although in classical instrumental conditioning there is no reference made to delayed reward processing. In RL terms instrumental conditioning can best be described as a multi-armed bandid problem.

[15] Of course guilt can arise in the absence of someone expressing anger. This interaction flow is presented as an example.

[16] This strongly relates to the hard problem of consciousness, qualia and whether or not consciousness is a natural phenomenon, i.e., a phenomenon that emerges from neurons just like molecules are an emergent phenomenon emerging from atoms.

**Emotions are Manifestations of Temporal Difference Processing**

The core of the TDRL Theory of Emotion is that all emotions are manifestations of temporal difference error. As argued above, the essence of an emotion is that there is positive or negative valence associated with it, that intensity and sign of valence is relative to what the individual is used to, and that emotions are involved in future behavioral change. Before discussing detailed evidence in favor of this proposal, I explain what this TD interpretation means for the emotions of joy, distress, hope and fear. I use these labels for particular emotions throughout this article. These labels refer to specific underlying temporal difference calculation processes and should not be taken literally or as a complete reference to all possible emotions that might exist in the categories that these labels typically refer to in psychology.

In this view, joy (distress when negative) refers to the positive (negative) temporal difference error (Figure 5). Based on an assessment of the values of different actions, an agent selects and executes one of those actions and arrives at a new situation. If this situation is such that the action deserves to be selected more (less) often, either because the reward was higher (lower) than expected or the situation is better (worse) than expected resulting in a positive (negative) temporal difference signal, the action value is updated accordingly. This update manifests itself as joy (distress). Joy and distress are therefore emotions that refer to the now, to actual situations and events.

The joy/distress signal - or in RL terms the temporal difference error - is also the building block for hope (fear). Hope (fear) refers to the *anticipation* of a positive (negative) temporal difference error (Figure 6). To explain fear and hope in the TDRL Theory of Emotion, the agent needs to have a mental model of the agent-environment interactions that is able to represent uncertainty and is used by the agent to anticipate. In RL this is referred to as model-based RL. In this case, the agent not only learns action value estimates (model-free RL, such as SARSA) but also learns probabilities associated with action-environment transitions $P(s'|s, a)$, i.e., what next state $s'$ to expect as a result of result action $a$ in state $s$. Fear and hope result from the agent's mental simulation of such transitions to potential future states. If the agent simulates transitions to next possible next states (either through its own actions in the environment or actions of other agents) and at some point a positive (negative) temporal difference error is triggered, then that agent knows that for this particular future there is also a positive (negative) adjustment needed for the current state/action. However, as this positive adjustment refers to a potential future transition, it doesn't feel exactly like joy (distress). It is similar, but not referring to the actual action that has been selected. How fear (hope) is incorporated in adjusting *current* action values to change behavior is not relevant here (I discuss this later). The point is that fear (hope) shares a basic ingredient with joy (distress), i.e., the temporal difference error assessment. There is a qualitative difference though, which has to do with the cognitive processing involved in generating the signal: while joy and distress are about the now, hope and fear are about anticipated futures.

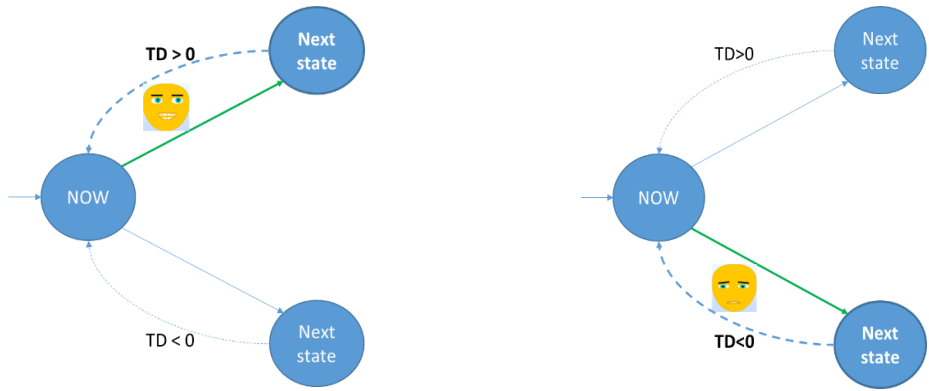

*Figure 5. Schematic for joy at the left (distress on the right). The agent / animal selects an action and ends up in a next state, resulting in a positive (negative) temporal difference error, which results in an increase (decrease) of the motivation for that action, which in instrumental conditioning is equivalent to a reward (punishment), resulting in the most basic positive (negative) emotion I would label joy (distress).*

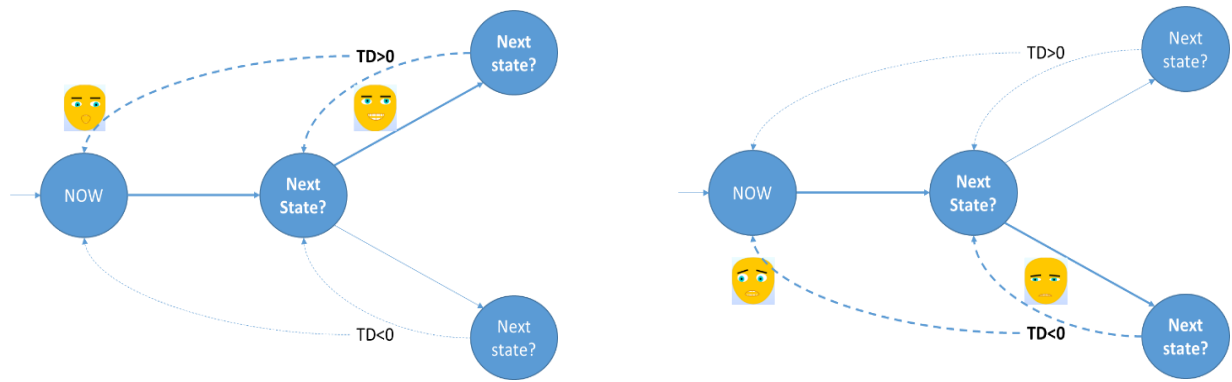

*Figure 6. Schematic for hope at the left (fear on the right). The animal **imagines** a possible future next state that when attained would trigger a positive (negative) temporal difference error, which would result in an increase (decrease) of the motivation for that action, which in instrumental conditioning is equivalent to the anticipation of a reward (punishment), resulting in a slightly more complex basic positive (negative) emotion I would label hope (fear).*

**Computational, Psychological and Neurobiological evidence**

In this section I discuss computational, psychological and neurobiological evidence supporting the TDRL Theory of Emotion. Evidence is grouped into three themes: similarities in the elicitation conditions of emotion and temporal difference errors, similarities in their functional effects, and similarities in their evolutionary and developmental purpose.

*The elicitation of emotion and the TD error is similar.*

With regards to emotion elicitation I discuss three important aspects: the situational circumstances that trigger the emotion, the valence of the emotion, and the intensity over time. I show that there are important similarities between emotion elicitation and TD error elicitation on these three aspects.

The vast majority of emotion theories propose that emotions arise from personally meaningful (imagined) changes in the current situation. An emotion occurs when a change is personally meaningful to the agent, i.e., when that change makes it more or less likely for the agent to result in a desirable situation. For example in cognitive appraisal theories the emotion is often described as a valenced reaction resulting from the assessment of personal relevance of an event (Moors et al., 2013; Ortony et al., 1988; Scherer, 2001). The assessment is based on what the agent believes to be true and what it aims to achieve as well as its perspective on what is desirable for others. In theories that emphasize biology, behavior, and evolutionary benefit (Frijda, 2004; Jaap Panksepp, 1998), the emotion is more directly related to action selection but the elicitation condition is similar: an assessment of harm versus benefit resulting in a serviceable habit aimed at adapting the behavior of the agent. Strongly cognitive views such as the one proposed by Reisenzein (Reisenzein, 2009b) assume that the mechanism by which this assessment is made is based on processes that check for congruency in the believes and goals the agent holds. For example, if the agent detects that a change in the current situation results in believing that a desirable (goal) state becomes true – i.e., the desired state and the current state are congruent - then that would trigger joy. Scherer (Scherer, 2001) proposes that the processes based on which this assessment is made can be split up into two: one based on biological congruence (intrinsic pleasantness) and one based on schematic congruence. Key is that in all these theories emotions are the result of a personally relevant change in the situation of the agent. Also in computational models that simulate emotions based on either cognitive appraisal theories (João Dias & Paiva, 2005; S. C. Marsella & Gratch, 2009; Popescu, Broekens, & Someren, 2014; Steunebrink, Dastani, & Meyer, 2008) or biological drives and internal motivation (Cañamero, 2003; Cos, Cañamero, Hayes, & Gillies, 2013) emotions always arise due to (internal) changes that are assessed as personally relevant.

The temporal difference error by definition arises exclusively from changes in the current situation and is related to personal relevance. TD errors arise if and only if (a) there is a state transition, and, (b) the transition generates a difference in expected return. The first means that only if there is a perceived change in the situation a TD error will arise. The second means that only if that change is assessed as personally meaningful - it increases or decreases the expected current and cumulative future reward – the TD error will be other than zero. In the TDRL theory of emotion, the desirability of a change is operationalized by the increase or decrease of the expected current and cumulative future payoff. This thus matches with the general elicitation conditions of an emotion: personally relevant and change driven.

There are important specific similarities in the elicitation conditions of emotion between TDRL emotion theory and other emotion theories. Most obviously it relates to the neurobiological theory of emotion by Rolls (Rolls, 2000), who proposes that emotions are the result of (absence of expected) reward and punishment. The TDRL theory differs in that it does not propose special roles for reward versus punishment for the elicitation of emotions and it generalizes the process of how anticipation-related emotions, such as fear and hope, arise. Further, TDRL emotion theory clearly separates between emotions proper (event-related valenced reactions of different complexity) and proto-emotions (including rage, sexual drive, pain, hunger). It also proposes how emotions of higher complexity develop due to cognitive complexity.

Further, also on the cognitive side there are important similarities in elicitation conditions. For example Reisenzein (Reisenzein, 2009b) proposes that there are two metacognitive processes of high importance for emotion elicitation: desirability (of a state) and likelihood (of that state to be true). The desirability of a state in Reisenzein's view directly matches with the TD error; desirable state are states you want to be *true* and states you want to be *true* are states with a positive TD error because those are the states that help you further optimize well-being. These states are in essence you goals. The likelihood matches with the probability that a particular state can be reached. There are two ways in which an agent can assess the likelihood of a state in model-based RL: first, the *current* state is certain by definition as it is actually happening now, second, the likelihood of future states can be assessed be querying a transition model in model-based RL. The emotions of joy and distress in Reisenzein's theory arise when a desired state has become certain (likelihood of 1). Reisenzein proposes this to be processes over symbolic representation, but this constraint is not needed for RL agents. For example, model-free RL, such as SARSA, can simulate joy and distress using the TD error in line with Reisenzein's proposal. For the emotions of hope and fear an anticipatory model of the environment is needed: to simulate hope and fear based on TD errors, we need model-based RL. For more detail on a potential mapping between RL and Reisenzein see (T. Moerland, Broekens, & Jonker, 2016). It is also possible to computationally formulate Scherer's appraisal processes in terms of RL primitives. Scherer (Scherer, 2001) proposes that the process of appraisal is split into different stimulus evaluation checks that operate at different levels of cognitive complexity. Again, the stimulus is a change in the current situation and the relevance, implications and whether or not the agent can cope with the change are then assessed. Scherer does not emphasize the labels of the emotions but rather focusses on the structure and flow of the appraisal process. First the relevance of the stimulus is assessed at different levels of complexity (suddenness, novelty, intrinsic pleasantness). This matches to the state change in RL and the associated reward signal strength. Almost immediately after that the implications are assessed. Most relevant for our current discussion is the assessment of conduciveness as this defines the desirability of the change. Conduciveness matches with the TD error. In (Jacobs, 2013) we propose a complete mapping between RL and Scherer's SEC's as well as the OCC model including an analysis of what these appraisal processes could mean in terms of RL-related learning signals. Although some of those mappings may later prove to be wrong, the essence is that for most appraisal processes (Ortony et al., 1988; Reisenzein, 2009b; Scherer, 2001) (except agency and norms related ones) an analogy can be found in terms of TDRL.

Another important observation is that the TDRL Theory of Emotion proposes a natural continuum towards more complex emotions based on the cognitive development of the individual or species. Joy/distress are manifestations of the TD error. For this there is no need for a model of the environment-agent interactions, i.e., no model-based RL. Hope/fear is modeled by anticipated TD error.

For this we need an anticipatory model (see more on this later in this section). To do this an agent needs to be able to imagine the future and assess the uncertainty of that future. The TDRL theory of emotion thus predicts that cognitively advanced species able to do so will most likely also experience fear- and hope-like emotions

The second aspect of emotion elicitation I discuss is emotional valence. The vast majority of emotion theories propose that an emotion is an evaluation of positive versus negative. This is true for cognitive appraisal theories (Moors et al., 2013; Ortony et al., 1988), theories emphasizing action and motivation (Bradley, Codispoti, Cuthbert, & Lang, 2001; Frijda, 2004), theories that focus on social evaluation (Fischer & Manstead, 2008), and theories that focus on basic survival needs and adaptive behavior (Jaap Panksepp, 1998; Rolls, 2000). In all these theories the emotional episode is an assessment of whether an (imagined) event is conducive, appetitive, pleasant (positive valence) versus obstructive, defensive, unpleasant (negative valence). Further supporting evidence for the fundamental role of valence is found in affective factor analysis of a wide variety of stimuli as well as verbal reports. A consistent finding is that the most explanatory factor for variance in emotions/words/attitudes and sentiments is *valence*, typically accounting for about 35% or more of the variance (J. R. Fontaine, K. R. Scherer, E. B. Roesch, & P. C. Ellsworth, 2007). Finally, in Social Signal Processing, valence is often found to be the best detectable social signal in facial expression, sentiment analysis and bodily behavior (Vinciarelli et al., 2012). Of course, valence is not enough to explain or describe all emotions, but valence is a fundamental aspect of emotional episodes. There is some debate (Macedo, Cardoso, Reisenzein, Lorini, & Castelfranchi, 2009; Ortony et al., 1988; Reisenzein, 1994, 2009a) on whether or not a neutral form of surprise should also be considered an emotion. Surprise could for example result from low-level assessment of novelty (Scherer, 2001), or, from the assessment of belief-belief congruence (Reisenzein, 2009b) i.e., a stimulus forces you to update your belief but there is no associated desirability with that belief). This type of surprise elicitation without valence is possible but in my view it is a definitional issue rather than a fundamental issue about the mechanism of emotion elicitation. "Neutral" surprise in nature seems to rare in the first place, and second, when it arises it mostly relates to information processing and sensory information uptake (vigilance, attention). In essence surprise is perhaps better defined as a state that a neural information processing system can be in (like sleep, flow and confusion).

The TD error in RL is always an assessment of positive versus negative. The TD error is positive (negative) if the state transition results in a higher (lower) than expected reward and cumulative future reward. This assessment is relative to the agent's "goals", where a goal in the TDRL sense is an anticipated future state with associated joy (i.e., a state you hope for). If a transition brings the agent closer to such as state, then the TDRL signal is positive and vice versa. If we simulate joy and distress as the TD error, then joy is elicited if the agent assesses the situation to be better than expected and vice versa for distress. (see Figure 5). The issue of whether or not neutral emotions exist (e.g. surprise) is dealt with by the fact that a TD can be zero but that would not result in any positive or negative valence. The event might still be unexpected (highly salient), and this can have behavioral and experiential consequences but is not related to "good or bad". Another issue, the correlation found in many studies between arousal and the absolute value of valence, e.g. in (J. R. J. Fontaine, K. R. Scherer, E. B. Roesch, & P. C. Ellsworth, 2007)), is also explained by the TD error interpretation. The more important the TD error (either positive or negative) is, the more likely the event was salient as well and the more likely action tendencies are changed. Highly salient events as well as changes in action tendencies are important aspects driving arousal through attentional resources and action readiness. Modelling joy and distress with the TD error therefore provides a natural explanation for the observed relation between arousal and valence.

Although it is tempting to equate the reward signal in RL itself to the positiveness / negativeness of the emotion, I argue, in line with e.g. (Murray, 2007), that this cannot be true for several reasons. Most of these are related to the fact that emotions are relative assessments with intensity dynamics that are more compatible with the TD error than the reward (see below). Two intuitive experiential aspects can be put forward too. First, rewards in nature are of specific types while emotions are more like a common motivational currency. In nature, reward is grounded in homeostasis / neurotransmitters and pleasure/pain centers. Reward is a multimodal primary reinforcer, it is about (dis)liking a stimulus for particular biological reasons, such as sugary food , gentle touch or pain. Emotions are not "multimodal feelings", you don't feel "money joy" versus "holiday joy", or, "pain fear" versus "loss fear". The positive/negative aspect of fear, joy, and all other emotions is a feeling of a single common "currency". Of course the agent still feels (anticipation of) pain or monetary loss as different things, but fear will be fear and joy will be joy. Second, emotions are experienced relative to the current situation. When you lose half of your one-million fortune you end up at the same amount of money as when you win your first half million. However, the feeling associated with these events is completely opposite. In the latter case, you could feel joy still owning half a million, taking the "no money" reference point, or, feel distressed about just losing the money taking the "one million" reference point. In the TDRL interpretation emotional valence and intensity depend on your TD error reference point. This aligns with Mellers' Affect Decision Theory explaining that unexpected outcomes have a higher emotional impact not due to the utility of the outcome but due to the unexpectedness (Mellers et al., 1997). Our TD theory of Emotion explains this because the TD update is larger in the case of unexpected outcomes, which can be simulated during the exploration phase of in computo learning task (Broekens, Jacobs, & Jonker, 2015). In line with findings on for example emotional coping, the emotional interpretation of an event (the perspective an individual takes during reappraisal) is rather flexible in human adults (Folkman & Lazarus, 1990).

I now turn to the third aspect of emotion elicitation: emotional intensity. The dynamics of the temporal difference error are compatible with the dynamics of joy and distress and provides a basis for the dynamics of fear and hope and even more complex emotions. To discuss emotional intensity, I focus on several key time-related emotion intensity modifiers: habituation, extinction, and regulation. Emotion habituation is the process of reacting less to repeated exposure of emotion eliciting stimuli, very similar to the process of habituation to reinforcers. Extinction is usually discussed in the context of fear extinction (Myers & Davis, 2006), and refers to the process of reacting less strong to repeated exposure to fear conditioned stimuli. In particular the process called *new learning* is relevant for our discussion. *New learning* explains fear extinction by proposing that novel associations with the previously fear-conditioned stimulus become more important after repeated presentation of that stimulus. This results in a decrease in fear response, not because the fear association is forgotten but because alternative outcomes become more important. Regulation, here, refers to the process of up- or downregulating emotional intensity of emotions due to the active imagination of a future situation. Depending on the details of the imagination process, emotions are either exaggerated or inhibited and the effect is different for imagined and actual emotions. I will show that the TDRL Theory of Emotion helps to explain habituation, extinction and regulation in a unified way.

If we adopt the view that emotions are adaptation-related feedback signals then emotions must be related to change, i.e., an emotion signals an important change. If there are no important changes (imagined or real), then emotions should be absent. If an agent has fully learnt to deal with its environmental challenges, then it does not need to change its behavior anymore and as a result does not receive any feedback signals to that end, hence it has no salient emotions[17]. This is observed in nature as well. Stable and predictable situations tend to be rather emotionless, while unstable, and unpredictable situations tend to elicit strong emotions. This is an indication that emotions are not related to the absolute value of the situation. One can win the lottery, and this elicits strong emotions, but there is no measurable difference in joy or even perceived quality of life one year after such an event (Brickman, Coates, & Janoff-Bulman, 1978). The same is true for negative events: people are very bad at predicting how they would feel a year after a major negative life event, and in general feel much better than they had predicted (Brickman et al., 1978). This leads me to conclude that the "absolute" value of states and actions in the RL mechanism are not a model for the emotions of joy or distress. Resultingly, the RL reward signal is also not a good model because value is the cumulative discounted future reward.

This discussion brings us to the first phenomenon, *habituation*. Repeated exposure to a rewarded stimulus (Figure 7) results in gradually less intense emotional reactions; we get used to it. Finally, when the reward is removed after habituation, the response is negative. The typical explanation is that we come to anticipate the response. In the TDRL interpretation, this means the reward got fully integrated in the state or action value due to repeated exposure, explaining the less intense response. When the reward is absent after habituation there is a negative TD and the state/value action is adjusted. This is distress. The TD error model explains emotional habituation. Rewards or state/action value does not. Both represent an objective quantity. Reward and values converge to stable values. A world full of rewards converges to a very high overall value for all of its states. This would feel very comfortable, but the joy associated with this world would definitely decrease over time.

In a similar manner, fear and hope cannot refer to the value of an action or state, even though it is tempting to propose that fear and hope are anticipations of high or low value (Broekens et al., 2015). If an agent learns to deal with a particular environment, the value of states/actions converge to a stable value. This is an arbitrary absolute value. In such a converged situation, the agent will not fear or hope things that are completely known and to which the agent is fully used to. What one fears (hopes) is distress (joy). Fear is thus anticipated distress, and hope is anticipated joy.

To explain fear (and hope) extinction in the TDRL Theory of Emotion, we need a model of the agent-environment interactions that is able to represent uncertainty and used by the agent to anticipate. In RL this is referred to as model-based RL. In this case, the agent learns a model of the action-environment statistics $P(s'|s, a)$, i.e., what next states $s'$ to expect as a result of an action $a$ in particular state $s$. As distress and joy are well modelled by the TD error, fear and hope are proposed to be anticipated TD errors using such an anticipatory model of the agent-environment relationship. Fear and hope are again signals that indicate important behavioral adaptation but now in the future. These signals habituate, as repeated exposure to predictable outcomes will result in the absence of TD errors. The only way to keep fear and hope existing is in the face of uncertainty, or, during learning. In the latter case (learning) the agent is still integrating rewards and learning the consequence of actions. If the agent comes to realize that a particular action causes distress in the future, it experiences fear at that same moment as it anticipates a negative TD (and vice versa for joy/hope/positive). In the former case (uncertainty), even if an agent has fully learned to act in an environment the agent can experience fear and hope if TD errors are stochastic. Walking along a dangerous mountain trail is frightening because you initially estimate the odds of slipping to be quite high. After a couple of hours walking, you have revisited these states many times and the odds are lowered, hence your feeling of fear is too. However, actively imagining the consequences of falling and the many ways in which you could indeed slip will enhance your fear experience. This also explains fear extinction by the process of new learning (Myers & Davis, 2006). Repeated visits of a previously fear conditioned stimulus result in lower fear because of new (more/heavier) connections with alternative outcomes. The fear is not forgotten but the fearful outcome is assessed as less likely. A more positive example is the buying of a lottery ticket that generates hope, not because one is learning and only realizes sitting behind the TV set that one might win, but because the outcome is uncertain and very positive. Anticipation of the potential joy (positive TD error) as a result of actual winning produces hope.

---

[17] This of course does not exclude the possibility that the agent experiences all kinds of other affective states due to remembering events, associations with stimuli, relations with other agents, moods, drives, etc.

This example bridges nicely to the phenomenon of *regulation*. When all your lottery numbers are correct except the last one, that last wrong number generates a lot of distress of a particular kind: *disappointment*. After every correct number you incorporate a bit of the positive TD error in your current state. That is also why you feel so joyful *just before* you realize the last number is incorrect. All these correct numbers generate small positive TD's, even though you know you are not rich yet, the odds of getting rich increase every time with a correct number. It doesn't matter if these are real odds or perceived odds. When the final number is shown and is not correct, you have to downgrade your situational value. In the TDRL Theory of Emotion, this is exactly what disappointment is, and it is in line with Reisenzein: the degree to which an anticipated amount of joy is in fact less than the actual amount of joy (Reisenzein, 2009b). This is again a TD error, accompanied by an even more complex cognitive state, a state that requires an anticipatory model and a memory enabling you to realize that you just negatively adjusted a value that was too positively adjusted based on previous anticipation (counterfactual thinking, imagination). Any animal able to imagine future rewards and punishments will experience hope and fear, but only those animals able to realize that the update was based on assumptions about the future will experience disappointment as a special case of distress (or relief as a special kind of joy). The rest simply experiences distress (with an expression that of course looks almost the same).

The last paragraph explains by example what *healthy* regulation does with the intensity of emotions. Regulation by anticipation brings the emotional event closer to the now, resulting in two important phenomena. First, the intensity of the expected emotion is spread out, because the TD error is already incorporated to some extent for the entire chain up until the emotional event. You feel a little bit of the event before it has happened. Second, when the event arises, it is less intense because some of the TD error has already been incorporated. Convincing yourself that you will fail an exam makes failing it slightly less bad. You feel the event but regulated. I propose several important behavioral functions for this in the next section. Also in the next section I explain what unhealthy anticipation does and why it does this from an RL point of view.

As a last note on habituation and regulation, but now on a longer time scale, consider the difference in emotion intensity between children and adults. The typically high-level explanation is that adults have a more developed prefrontal cortex, that the prefrontal cortex is involved in inhibition, regulation and control, and hence children are less able to control their emotions. The TDRL Theory of Emotion proposes an alternative explanation. Adults are converged agents, they are relatively finished in terms of learning and therefore rarely experience strong TD errors. Adults are "stable" in the sense that they have a fully developed model of how they interact with the world and what that means in terms of rewards. As such, it is to be expected that adults experience less intense and less frequent emotions compared to children, for two reasons: most of their behavior is built out of habits, and, they are able to anticipate and therefore spread out large fluctuations in TD errors. This is compatible with the more recently discovered functions of the prefrontal cortex involving value and TD error calculations (Berridge et al., 2009; Haruno & Kawato, 2006; Suri, 2002; Tanaka et al., 2004). It is also compatible with the view propose earlier that emotions are a natural consequence of the need for ever more intricate adaptation of reflexes shaping complex behavioral responses to help survival of the individual. If such behavioral feedback signals are less frequent, emotions should be less frequent too.

In this section I showed how the TDRL Theory of Emotion can be used to understand the elicitation conditions, emotional valence and intensity of emotions. I explained how the TD error models joy and distress, and how form that hope, fear, disappointment and relief emerge given more cognitive complexity. I proposed a TD error explanation for the effect of habituation, fear extinction and anticipation on emotion intensity.

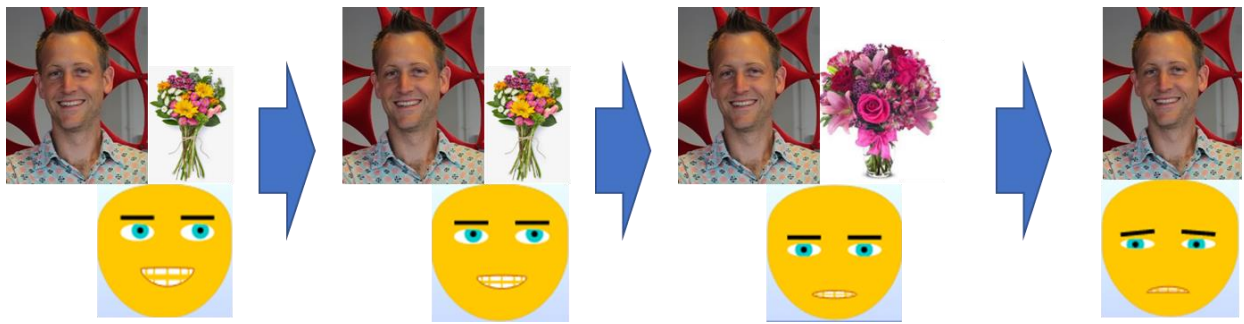

*Figure 7. Graphical schema of habituation of joy due to the repeated result of giving flowers to someone (iconic face), ending in distress when the rewarding stimulus is not presented after habituation.*

*The functional effect on behavior of emotion and the TD error is similar*

Emotions are tied to motivation, action and action tendencies (Frijda, 2004; Jaap Panksepp, 1998). Specific behaviors are often proposed to be related to specific emotions: fear and fleeing, anger and fighting, joy and play, sexual drive and mating behavior, to name just a few. Emotions are also tied to behavioral feedback, reflection and anticipation (Baumeister et al., 2007). Finally, emotions are essential for decision making and action (Damasio, 1994). The difference in these views is not about the ultimate function of emotion, i.e., to influence future action, but about what comes first: does the emotion cause the action, does the emotion provide feedback about the action, does the emotion cause bodily changes, do the bodily changes cause an emotion. I propose that the TDRL Theory of Emotion sheds light on how this process works.

In Reinforcement Learning the motivation for action is encoded as action value. RL essentially is about learning to select appropriate actions. Appropriate means "optimizing cumulative future reward". The action value in RL thus models motivational value. There is evidence that RL-like motivational action value is also encoded in the brain (Berridge & Robinson, 2003; Tanaka et al., 2004). The value of an action in RL is a direct driver for the selection of that action, computationally represented as some form of probabilistic action selection rule (*Formula 2*). This probabilistic selection can vary from begin very random, read exploratory behavior, to very selective, read goal oriented.

Action values are incrementally updated by the temporal difference error. The TD error is an adjustment of the action value based on new evidence after having executed an action (*Formula 1*). There is also evidence that TD errors are encoded in the prefrontal cortex (Haruno & Kawato, 2006; Suri, 2002).

$$p(a) = \frac{Q(s,a)}{\sum_{i=1}^{|A|} Q(s,a_i)}$$

[2]

If emotions are the manifestations of TD errors, then this explains why emotions are related to motivation, motivational change, feedback on actions, and action readiness all at the same time. The neural calculation of the TD error drives the motivational change as it updates the action value. It thereby functions as a feedback signal that changes the action value. Any action that becomes strongly motivated must also be emotionally relevant because this involves an important TD error. Especially when an agent has an anticipatory model (as explained above), this interpretation makes sense. Consider a hungry agent in a state of choice. One option leads to nothing particular, the other option leads to a great meal. If the agent is able to anticipate this meal to some extent, then, following our model of hope/joy, the agent will propagate back some of the anticipated joy to the current state of choice. This must also increase the motivational value of the option leading to the meal. It also makes the agent feel hopeful (again, hope is just a label here). As a result the agent prepares for the approaching action and starts committing to it. This process is easily simulated in model-based RL (T. Moerland et al., 2016). This shows that the anticipation of a positive TD can alter current action motivation and produce feelings of positive anticipation (to avoid the label of hope for once). This explanation obviously also holds for fear and the avoidance of distress, and it neatly ties together emotion, motivational change and action. It also provides a natural continuum for the three types of Baumeister "feedback" (Baumeister et al., 2007): feedback, anticipation and reflection. All three are essentially the same type of feedback signal, but executed at different time scales and hence in need of increasingly complex cognitive capabilities. Feedback is about joy and distress and "is" the TD error. The anticipation emotions hope and fear are about anticipated joy and distress, but as joy and distress are feedback signals, both hope and fear are also feedback, but they are feedback about an anticipated future. This ties hope and fear strongly to motivation. Reflective emotions such as disappointment (relief) are about the difference between anticipated joy and actual joy (distress). Again, the signal is a TD error. The reflection part is due to the realization that the TD error could have been different. This requires even more cognitive sophistication. Although Baumeister and Reisenzein formulate their emotions differently, the TDRL Theory proposes that they are in agreement about what emotions are.

We see that the TDRL Theory of Emotion is able to integrate these different theoretical views *and* ties emotions to a computational method that is functionally able to learn and develop highly complex tasks as well as is grounded in neurobiology. The TDRL Theory of Emotion can also explain and predict a variety of cognitive-affective phenomena, some of which are functional while others are maladaptive. I will proceed to highlight several.

A well-known phenomenon in psychology is that strong anticipations of positive or negative outcomes create a reversed effect when the anticipated event does not follow through or has only moderately rewarding or punishing effects. The stronger the anticipation of gain (loss) the stronger the disappointment (relief) (Reisenzein, 2009b). The TDRL Theory explains this as follows. If you simulate (anticipate) a trace towards a joyful future, you feel the positive TD error, but also incorporate a little bit of it in the trace already, which is necessary for action motivation towards that "goal", as explained above. By the time you arrive several things can happen. If you underestimated the gain, you will still feel joy because the TD is positive. If you overestimated the gain, you will feel a particular form of distress I label as *disappointment* because the TD is negative due to the needed negative adjustment.

*Healthy* anticipation thus has an adaptive function for at least two reasons: it drives motivation and goal oriented behavior, and, it regulates extreme TD errors by smoothing over the anticipated trace. This is useful in both the positive (hope) and negative case (fear).

*Unhealthy* positive anticipation is characterized by strong positive TD errors based on an anticipated future resulting in elation and regular disappointment (in the hope case), and, strong negative TD errors based on an anticipated future resulting in a lot of distress (in the fear case). This mechanism is an interesting candidate for an explanation of bipolar disorder (manic depression) characterized by alternating optimistic versus pessimistic modes of at least a week[18]. If you flip-flop between incorporating very positive TD errors on the one hand and very negative TD errors on the other, than the observed behavior (also "in computo") is extreme motivation and joy while going for high gains with small odds followed by strong disappointment, flipping to inactivity, distress and worrying explained by the constant anticipation of negative TD errors for all possible actions. The TDRL Theory of Emotion thus suggests that bipolar disorder is a stability disorder of TD error anticipation and regulation.

We performed in computo simulation of how the intensity of fear is modulated by perception of control, amount and depth of rumination and closeness of threat (T. Moerland et al., 2016). Individuals with high perception of control believe that they control their own actions, that their actions will not fail and that those actions will have a predictable, usually deterministic, outcome. Individuals with a high perception of control typically are less fearful and take more risk (Horswill & McKenna, 1999). In our lab we manipulated in computo experiments the amount of control individuals assumed they had when anticipating walking along a slippery cliff. Individuals with higher perception of control experienced less fear than individuals with lower perception of control. Perception of control was operationalized as the mental action selection strategy (more versus less randomness in action selection). Rumination involves repeated mental simulation of possible future interactions with the environment. High levels of rumination produced higher levels of fear. We also simulated the amount of internally simulated traces while walking along the cliff. More simulation again produced higher fear intensity. Closeness of threat refers to the temporal closeness of harm. Simulation results showed that when the cliff closer, fear intensity is higher. These studies show that when fear is simulated as anticipated future TD error signals, we are able to replicate in computo three important findings from psychology.

Finally we address the correlation between sensation-seeking, risk-taking behaviour, pathological gambling, and (perception of) control over event outcomes (Anderson & Galinsky, 2006; Fortune & Goodie, 2010; Horswill & McKenna, 1999; Nower, Derevensky, & Gupta, 2004; Strickland, Lewicki, &

---

[18] https://online.epocrates.com/diseases/48836/Bipolar-disorder-in-adults/Diagnostic-Criteria

Katz, 1966). Explanations for this co-occurrence include impulsiveness of the individual (Potenza, 2008; Romer et al., 2009), perceived self-efficacy (Krueger & Dickson, 1994), and optimistic risk perception. Optimistic risk perception results from event outcome probability bias (Anderson & Galinsky, 2006; Krueger & Dickson, 1994) characterized by higher regard for positive outcome probabilities compared to negative ones. In computational simulation studies (Broekens & Baarslag, 2014) we showed that intensity of sensations, risk-taking, pathological gambling, and lower levels of fear co-occur when optimistic risk perception underlies the temporal difference (TD) signal used in learning to adapt to a risky task. We found that optimistic calculation of the TD signal, compared to different forms of realistic calculation, results in risk taking and in persistent gambling behavior. We also found that an optimistic TD signal results in less fear and more intense joy. We found these results to be consistent for a variety of risky tasks. This computationally replicates the finding that effects of perception of control on risk-taking are mediated by optimistic risk perceptions (Anderson & Galinsky, 2006) and replicates the positive correlation between sensation intensity and pathological gambling (Powell, Hardoon, Derevensky, & Gupta, 1999). In that work we argue that the simulations underline the importance of the relation between the prefrontal cortex (as a locus of control and outcome value representation) (Kennerley, Behrens, & Wallis, 2011; Rolls & Grabenhorst, 2008; Romer et al., 2009; Rushworth & Behrens, 2008), the amygdala (as a locus of gated value recalculation and fear processing) (Bechara et al., 2000; Murray, 2007; Seymour & Dolan, 2008) and gives further evidence for Reinforcement Learning being a computational model for reward processing, emotion and task-adaptive behavior (Berridge et al., 2009; Dayan & Balleine, 2002; Murray, 2007; O'Doherty, 2004; Redish et al., 2007; Rolls, 2000; Rolls & Grabenhorst, 2008; Schultz et al., 1997; Suri, 2002).

*The evolutionary "purpose" of emotion and the TD error is similar*

From an evolutionary perspective emotions serve behavioral adaptation of the individual. Emotions aid in action selection by promoting activity (Frijda, 2004), emotions aid in assessment of the situation (Damasio, 1994; Ellsworth & Scherer, 2003), emotions are internal feedback signals changing future behavior (Baumeister et al., 2007). And finally, emotions are important social signals (Paul Ekman & Friesen, 2003). Given such strong involvement of emotions in action, many emotions are probably not tied to humans alone but widespread in nature. To fully understand the phenomenon of emotion it needs to be investigated cross species (Bekoff, 2008; J. Panksepp, 1982; Jaap Panksepp, 1998) .

What is the purpose of emotion for the individual animal? Building on this cross-species view, I propose that emotions are short-term mechanisms to aid an agent's long term survival and as a byproduct optimize long term well-being of the agent. This view of emotion is not new or unique to the TDRL proposal. For example, Cabanac (Cabanac, 1992) already proposed that one of the first mental events to have emerged in nature is the ability to experience pleasure and displeasure in order to optimize survival. Here, however, it is important to understand TDRL and emotion in relation to survival of an agent. The feeling of overall happiness (well-being) is relatively stable against important life events in humans (Brickman et al., 1978; Veenhoven, 1991). However, when important social or biological needs are repeatedly not met, overall well-being is definitely impacted, as is the case in human poverty, but also stimuli- and socially deprived animals. More food, friends, money or power do not make an individual necessarily happier in the long run, but a severe lack of such means does negatively impact well-being. This points to the view that there is a threshold that defines "good enough". Emotions aid the individual in making those decisions that work towards this threshold. They help us to decide what to do next, relative to how we are currently doing. If you have a lot of friends, but no food, then perhaps it is a good idea to focus on getting some food. Evolutionary speaking, *complex organisms need to have a motivational heuristic for a potential next action, relative to how the whole organism is doing as well as relative to alternative actions*. This is in line with *causal factor space* in which it is proposed that the motivational force for behavioral alternatives is assessed relatively (McFarland & Sibly, 1975). This view is directly advocated by Damasio, who proposes somatic markers as a construct for this heuristic (Damasio, 1994), and resonates perfectly with TDRL as explained next.

The computational view of TDRL aims at optimizing utility expressed as the cumulative discounted future reward. Optimization in nature can be relaxed to *satisficing*, trying to achieve a level that is good enough for survival, which mirrors the "good enough threshold" of well-being. If reward and utility are grounded in multimodal homeostasis related to pain, food, etc., as proposed earlier and compatible drive-based rewards in RL literature, then TDRL perfectly models well-being, at least to a level that is good enough to survive comfortably. As utility is directly represented in the values of actions, changes in utility due to actions generate TD errors and should therefore also generate emotions. *The purpose of TD errors and emotions are therefore the same: to signal significant changes in the well-being of the agent.*

Finally, in many species emotions also have a social function. Emotions are expressed and recognized (Bekoff, 2008; Kret, Jaasma, Bionda, & Wijnen, 2016). Functionally, emotional expressions are social signals aimed at regulating (or perhaps influencing is a better term here) the behavior of others. For such a social signaling mechanism to be beneficial to the individual and species, there must be agreement on what such signals mean. This shared understanding can only evolve if there is evolutionary pressure on reproduction. For the expression and recognition of many emotions this evolutionary pressure is straightforward and argued already by many others[19] (Bekoff, 2008; P. Ekman & Friesen, 1971; Jaap Panksepp, 1998). The TDRL theory of emotion explains that the driving force for this social signal is the sender's individual TD error. I give three simple examples of how an individual TDRL signal can trigger the generation of a social signal. First, rather than physically punishing another individual for actions that harmed you, your expression of anger helps avoiding physical harm to you and the person you are angry at and serves the same purpose: *the expression of anger is a social signal aimed at communicating a negative TD*. Second, rather than feeling miserable and die, the expression of sadness will help an infant to get appropriate attention from her parents: *the expression of sadness is a social signal aimed at communicating the need for help*. Third, the expression and recognition of guilt assures that the angry other knows *you* incorporated their scorn in your future behavioral tendencies. This avoids the built up of aggression and retribution: *the expression of guilt is a social signal aimed at communicating the incorporation of a (socially) communicated negative TD*. I could go on, but in all of these cases, the emotional expressions have to do with the signaling of experienced TD's (anger, sadness, guilt) and the confirmation that the signal has been correctly received (guilt, worry, forgiveness). The more complex the signal the less likely a particular expressive pattern for that TD error evolved. This explains why joy, sadness and anger are among the best recognizable emotions, even cross species. They are related to immediate TD effects and expression is close to the typical behaviors accompanying such mental states (play and laughter, immobility and crying, fighting and shouting). They are extremely important for conflict resolution and child raising. The TDRL theory of emotion is therefore compatible with the view that emotions are social phenomenon, and proposes that emotions are widespread in animals that raise children or have the need to resolve social conflict.

---

[19] Although the number of facially expressed emotions that are cross culturally recognized by humans is a currently debated topic, there is no doubt that the multimodal expression and recognition of some emotions, such as joy and sadness, is universal even across different species of mammals (Bekoff, 2008).

**Discussion of limitations and ramification of the approach**

We showed that the TDRL Theory of Emotion is compatible with established cognitive explanations of emotion such as appraisal theory. In addition to these cognitive views, the TDRL Theory of Emotion is also able to explain how emotions relate to both task learning and cognitive processing as well as grounds the emotion in adaptive behavior. It also provides a clear and simple explanation for why all emotions have associated positive or negative valence, and are related to action tendencies. It also proposes why emotions are different from but correlated with hedonic pleasure (liking: the multimodal reward signal). Finally, it explains the emergence of more complex emotions as a result of an agent's cognitive capabilities such as anticipation, internal simulation of interaction, and agency detection.

It is important to also determine the boundaries of the TDRL theory of Emotion. The theory does not provide a clear explanation for affective reactions resulting from music and humor. Although it is tempting to propose that musical harmony, like play, is an inherently positive stimulus forming the basis for positive reactions to music, I do not go this far. Similarly one could propose that humor has to do with the unexpected positive outcome of a verbal puzzle. However, I believe affective phenomena involved in humor and music are at this point out of bounds for this theory.

The TDRL Theory of Emotion in its current form does not address mood. Mood is the longer term undifferentiated, less intense form of affect (Beedie et al., 2005). It has been shown that mood primes mood congruent memories (Bower, 1981) and attention towards congruent stimuli (MacLeod, Mathews, & Tata, 1986). However, the relation between mood and emotion is far from clear. For example, depression is associated with reduced emotional reactivity to sad contexts (Rottenberg, 2005). A possible interpretation of mood in terms of TDRL is that mood is the current feeling of well-being (the value or utility of the organism) while emotions are the TD errors. This would explain the previous finding: if the current situation is already quite negative, then additional negativity will not trigger high TD errors because neurobiologically it is impossible to end up much worse, therefore the emotional reaction is also reduced. However, this interpretation needs further investigation.

Social emotions such as anger and guilt have been explained as feedback signals above, but such emotions do need more than standard model-free or model-based RL. To explain these, an agent needs the capacity to attribute agency to the actions of another agent. The typical RL framework does not address this. If RL is used in a multi-agent setting, in which signals and behavior can be detected to be originating from other agents, then this is a start to better explain social emotions. This is important future work.

I proposed the labels joy, distress, hope and fear for the positive, negative and anticipated positive and negative TD error respectively. Also I proposed that disappointment and relief refer to the correction to the anticipated TD signal, with relief being a positive correction for an anticipated negative TD and disappointment a negative correction to an anticipated positive TD. Other labels could be proposed (such as anger and guilt, as mentioned earlier) but the question remains why one would label these RL-related signals as such. Answering this question involves the following steps, First, one should analyze theoretical consistency. For example, model-free RL does not contain a model that can be used for anticipation, and hop and fear need anticipation, therefore emotions resulting from anticipation must assume some model capable of doing forward simulation. So, labels for anticipation emotions that do not involve a form of look ahead in the RL model would not pass this test. Second, one should replicate generic principles of emotions during learning tasks *in computo* such as fear extinction and joy habituation. Third, one should replicate *data* on emotional intensities found during experiments. In this and previous articles (Broekens et al., 2015; T. Moerland et al., 2016). we have shown theoretical consistency and *in computo* replication of overall principles. Data replication studies are important future work.

Human adults are noisy subjects to study emotion. To study the basis of emotion and affect, human adults are not suitable for the following reasons. First, adults have high level of control over the sensory influence on the content of their working memory. Human adults are able to internally simulate fictional interaction with their environment (thought as internal simulation of behavior (Cotterill, 2001; Hesslow, 2002)) as well as have a hard time to stop anticipating (stay in the now). This enables them to imagine uncontrollably during an experiment influencing any learning or emotion task generating new emotions as well as up or downregulating emotional intensity. Second, human adults have developed strong coping and reflection capabilities (emotion down regulation). As a result, emotions will not be expressed or even felt with the same intensity they are generated in young children or animals. Third, affective associations have been built up for decades (attitudes producing emotions due to association). As a result, no emotion is the result of the appraisal of a stimulus or though alone but also of a large set of associated attitudes that are personal and uncontrollable in an experiment. Fourth, adults of any species have formed a rich set of habits, and habits will downregulate emotions (e.g. through extinction and joy habituation). As a result, the signal you try to measure is weak at the basis already, especially in controlled experiments as this is such a common setting in general for many people (office, desk, pushing computer buttons, etc…). Fifth, language plays an important role in the labeling of felt emotions, and this labeling process probably strongly influence how adults think about emotions as well as classify emotions. As such there is probably important interplay between emotion regulation, coping and language use. These reasons lead me to conclude that to study the basis of emotion, the human adult condition is unsuitable and it is arguably an unusual condition from an evolutionary point of view. This highlights how difficult it is to study the fundamentals of emotion in human adults. Perspective taking, coping, reference points as well as elaborate use of language are cognitively highly advanced capabilities that are hardly representative of the fundamentals of emotional processing in nature. *Concludingly, it is essential that future research focusses on psychometric instruments that are capable of assessing the experienced emotional intensity in animals and children in a reliable and valid way.*

*Relevance to the advancement of human robot/agent interaction.*

The diversity of tasks and settings in which robots and virtual agents need to operate prohibits preprogramming all possible behaviors in advance. Such agents need to learn from interaction with humans (W. Bradley Knox, Stone, & Breazeal, 2013; Thomaz & Breazeal, 2006). Machine learning needs a human in the loop. Humans use emotions as a natural way to communicate and emotional communication is essential in the learning process of infants (Buss & Kiel, 2004; Chong et al., 2003; Klinnert, 1984; Trevarthen, 1984). Typically the aim in this field of developmental robotics is that robots develop task-related skills in similar ways as children do: by trial and error, and by expressing emotions for help and confirmation, and learning from emotions expressed by others for shaping their behavior.

Reinforcement Learning (RL) (Sutton & Barto, 1998; Tesauro, 1995) is a well-established computational technique enabling robots and agents to learn skills by trial and error, for example learning to walk (Kober et al., 2013). Also, RL can deal with large state spaces when coupled to pattern detection using deep-learning (Mnih et al., 2013).

At this point there is no generally accepted computational framework that links emotions to reinforcement learning (see (T. M. Moerland, Broekens, & Jonker, 2018) for a review). This means that currently robots and virtual characters that learn based on RL lack the ability to produce and interpret emotions in line with their learning process. There are frameworks based on cognitive appraisal theory (Joao Dias, Mascarenhas, & Paiva, 2011; S. Marsella, Gratch, & Petta, 2010; S. C. Marsella & Gratch, 2009), but these models assume that emotions arise from a cognitive reasoning process, not a learning process based on exploration and positive and negative feedback. There is also research showing that robots and agents can use human feedback as signals to influence their learning process based on reinforcement learning (Broekens, 2007; W. Bradley Knox, Glass, Love, Maddox, & Stone, 2012; W.B. Knox & Stone, 2010; W. Bradley Knox et al., 2013; Thomaz & Breazeal, 2006; Thrun et al., 1999), and this provides a good starting from a human in the loop learning perspective.

A computational model of emotions based on the TDRL Theory of emotion will enhance interactive learning abilities of robots and virtual agents. It will make the learning process more *explainable* and *transparent* for the users of that robot. Emotions that are simulated and expressed by the robot are grounded in the learning process of the robot. Emotional expressions of the human user can be properly interpreted by the robot in terms of learning-related signals. (fig 1).

Concludingly, the development of a TDRL Theory of emotion is instrumental for the embedding of smart social robots and virtual agents in our society.

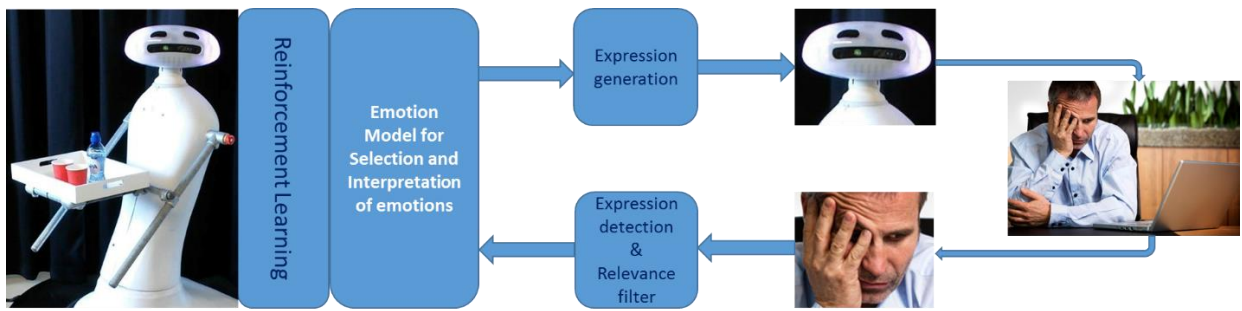

*Figure 1. A robot learns a task using RL as its controller. The computational model of emotion selects an appropriate emotion to express based on the current state of the learning process. The robot expresses the emotion. The robot owner interprets the signal and situation, and reacts emotionally to the robot. The robot detects and filters the reaction for relevance to the task (e.g., the owner's frown could also be due to reading a difficult email). The computational model of emotion interprets the owner's emotional reaction as learning-related feedback.*

*Emotions and animal welfare*

There is abundant evidence showing that other animals are emotional beings too (Bekoff, 2008). Recent evidence includes bonobos showing attentional bias towards pictures of emotional bonobos as compared to neutral bonobo pictures (Kret et al., 2016), social-relation driven reciprocity of agnostic support (a bystander helping one individual in conflict because you like him/her) in ravens (Fraser & Bugnyar, 2012), and separation anxiety in dogs and cats (Schwartz, 2003). To quote Bekoff (Bekoff, 2008), "it is bad biology to propose that animals do not have emotions", for very straightforward reasons: animal brains are structured like ours, their behavior looks like ours, they respond the same to many of our drugs, and their evolutionary roots are the same as ours. Why would they completely lack emotionality if it so important for humans?

However, for some species, most biologists would agree that they do not possess an internal mental life and therefore probably no emotions either. For example, flies, spiders, and zebrafish, would probably not feel or care about anything in our sense of the words caring and feeling, while bonobos, rats, dogs and raven do. So it seems that at some point it makes no sense to talk about emotions.

I propose that the TDRL theory of emotion draws a useful "line". I propose that naturally evolved species capable of incorporating delayed motivational effects of actions in their future action selection process must also possess some rudimentary form of emotion. The reasoning for this is that to be able to incorporate such delayed motivational effects (reward and punishment) they need to be able to assess the temporal difference signal associated with the effect. This signal is the animal's assessment of change in well-being and hence is a sufficient condition for a rudimentary form of joy versus distress.

This breaks up the world of living beings into those that show temporal difference learning and those that do not. It provides a clear cut, and easily studied criterion: whenever a species shows adaptive behavior that can only be explained with TD error processing, it must be considered as a candidate for emotions as well. As already argued above, quite a lot of species would already fall within the set of emotional beings. All animals that show parenting, play and social conflict resolution, telltale signs of reflex shaping through (social) TD error processing, must be considered emotional in the proper human sense.

**Conclusion**

I proposed that all emotions are manifestations of Temporal Difference (TD) error assessment. I discussed computational, psychological and neurobiological evidence. I explained how the TDRL Theory of Emotion explains the links between emotion, action, motivation and appraisal. I discussed the ramifications of this proposal for emotions as social signals, emotions in animals, and the relation between emotion, cognition and task learning. I conclude that the TDRL Theory of emotion can explain a wide variety of emotion-related phenomena in a unified way, and proposes many lines of future research.

**Acknowledgements**

Running the risk of forgetting someone, I would nonetheless like to thank several people who, over the years, have helped me shape this idea. In loose order of appearance these are: my Affective Computing class students for working on earlier versions of this idea; Stacy Marsella for co-organizing the 2011 Lorentz workshop on emotion modelling and inviting me to talk at USC on this topic; Elmer Jacobs for doing his Master thesis on analyzing the relation between RL-emotion and Appraisal Theory; Catholijn Jonker for always supporting me in this line of research; Agneta Fisher, Lola Canamero and other participants to the 2016 Lorentz workshop on emotion as feedback signals for emphasizing that RL and reward are not the only things that matters for emotion; Thomas Moerland for extending the initial ideas and testing them in model-based RL; Heather Urry for inviting me for a flash talk on the topic at SAS 2017 enabling me to meet and talk with Luke Chang; and last but not least Jon Gratch for his insights and detailed feedback on an earlier draft.